\documentclass{llncs}
\usepackage{epsfig,amsmath,amsfonts,latexsym,algorithm,algorithmic}

\newcommand{\PPP}{\mathcal{P}}				

\newcommand{\keywords}[1]{\par\addvspace\baselineskip
\noindent\keywordname\enspace\ignorespaces#1}

\begin{document}

\title{Online prediction of ovarian cancer}

\author{Fedor Zhdanov, Vladimir Vovk, Brian Burford, Dmitry Devetyarov, Ilia Nouretdinov and
Alex Gammerman}
\institute{Computer Learning Research Centre,
  Department of Computer Science\\
  Royal Holloway, University of London,
  Egham, Surrey TW20 0EX, UK\\
\email{fedor@cs.rhul.ac.uk}
}

\maketitle

\begin{abstract}
In this paper we apply computer learning methods to diagnosing ovarian cancer
using the level of the standard biomarker CA125
in conjunction with information provided by mass-spectrometry.
We are working with a new data set collected over a period of 7 years.
Using the level of CA125 and mass-spectrometry peaks,
our algorithm gives probability predictions for the disease.
To estimate classification accuracy
we convert probability predictions into strict predictions.
Our algorithm makes fewer errors
than almost any linear combination of the CA125 level and one peak's intensity
(taken on the log scale).
To check the power of our algorithm
we use it to test the hypothesis that CA125 and the peaks
do not contain useful information for the prediction of the disease
at a particular time before the diagnosis.
Our algorithm produces $p$-values
that are better than those produced by the algorithm
that has been previously applied to this data set.
Our conclusion is that the proposed algorithm
is more reliable for prediction on new data.
\keywords{Online prediction, aggregating algorithm, ovarian cancer, mass-spectrometry, proteomics}
\end{abstract}

\section{Introduction}
Early detection of ovarian cancer is important since clinical symptoms sometimes do not appear until the late stage of the disease. This leads to difficulties in treatment of the patient. Using the antigen CA125 significantly improves the quality of diagnosis. However, CA125 becomes less reliable at early stages and sometimes elevates too late to make use of it. Our goal is to investigate whether existing methods of online prediction can improve the quality of the detection of the disease and to demonstrate that the information contained in mass spectra is useful for ovarian cancer diagnosis in the early stages of the disease. We refer to the \emph{combination} of CA125 and peak intensity meaning the decision rule in the form
\begin{equation*}
u(v,w,p) = v \ln C + w\ln I_p,
\end{equation*}
where $C$ is the level of CA125, $I_p$ is the intensity of the $p$-th peak, and $v,w$ are taken from the sets described below.

We consider prediction in \emph{triplets}:
each case sample is accompanied by two samples from healthy individuals,
\emph{matched controls},
which are chosen to be as close as possible to the case sample
with respect to attributes such as age, storage conditions, and serum processing.
In the given triplet of samples of different individuals we detect one sample which we predict as cancer. This framework was first described in \cite{Gammerman2008}. The authors analyze an ovarian cancer data set and show that the information contained in mass-spectrometry peaks can help to provide more precise
and reliable predictions of the diseased patient than the CA125 criteria by itself
some months before the moment of the diagnosis. In this paper we use the same framework and set of decision rules (CA125 combined with peak intensity) to derive an algorithm which performs better
in some sense than any of these rules.

For our research we use a different more recent ovarian cancer data set \cite{Menon2005} processed by the authors of \cite{Devetyarov2009} with a larger number of items than in \cite{Gammerman2008}. We combine decision rules proposed in \cite{Devetyarov2009} by using an online prediction algorithm\footnote[1]{A survey of online prediction can be found in \cite{CesaBianchi2006}.} and thus get our own decision rule. In
this paper we use a combining algorithm described in \cite{VovkPEABG}, because it allows us
to output a probability measure on a given triplet and has the best theoretical guarantees for this type of prediction. In order to estimate classification accuracy, we convert probability predictions
into strict predictions by the \emph{maximum rule}: we assign weight 1 to the labels with maximum predicted probability, weight 0 to the labels of other samples, and then normalize the assigned weights.

We show that our algorithm gives more reliable predictions
than the vast majority of particular combinations
(in fact, more thorough experiments, not described here,
show that it outperforms all particular combinations).
It performs well on different stages of disease.
And when testing the hypothesis that CA125 and peaks
do not contain useful information for the prediction of the disease
at its early stages,
our algorithm gives better $p$-values
in comparison to the algorithm which chooses the best combination;
in addition, our algorithm requires fewer adjustments.

Our paper is organized as follows. In Section~\ref{sec:frame} we describe methods we use to give predictions. Section~\ref{sec:data} gives a short description of the
data set on which we work.
We show our experiments and results in Section~\ref{sec:experiments},
separated into description of the probability prediction algorithm in Subsection~\ref{ssec:alltr} and detection at different stages before diagnosis in Subsection~\ref{ssec:early}.
Section~\ref{sec:conclusion} concludes our paper.

\section{Online prediction framework and Aggregating Algorithm}\label{sec:frame}
The mathematical framework used in this paper is called prediction with expert advice. In this framework different experts predict a sequence of events step by step. The ones that make errors suffer loss defined by a chosen loss function. The goal of an online prediction algorithm is to combine the experts' predictions in such a way that at each step the algorithm's cumulative loss is close to the cumulative loss of the best expert. Unlike statistical learning theory, online prediction does not impose any restrictions on the data generating process.

A game of prediction consists of three components: the space of outcomes $\Omega$, the space of predictions $\Gamma$, and the loss function $\lambda: \Omega \times \Gamma \to \mathbb{R}$, which measures the quality of predictions. In our experiments we are interested in the \emph{Brier game} \cite{Brier1950}, since it is widely used in probability forecasting.

Let $\Omega$ be a finite and non-empty set,
$\Gamma:=\PPP(\Omega)$ be the set of all probability measures on $\Omega$.
The Brier loss function is defined by
\begin{equation}
  \lambda(\omega,\gamma)
  =
  \sum_{o\in\Omega}
  \left(
    \gamma\{o\} - \delta_{\omega}\{o\}
  \right)^2.
\end{equation}
Here $\gamma \in \Gamma$ and $\delta_{\omega}\in\PPP(\Omega)$ is the probability measure
concentrated at $\omega$:
$\delta_{\omega}\{\omega\}=1$
and $\delta_{\omega}\{o\}=0$ for $o\ne\omega$.
For example, if $\Omega=\{1,2,3\}$, $\omega=1$,
$\gamma\{1\}=1/2$, $\gamma\{2\}=1/4$, and $\gamma\{3\}=1/4$, then
$\lambda(\omega,\gamma)=(1/2-1)^2+(1/4-0)^2+(1/4-0)^2=3/8$.

The game of prediction is being played repeatedly
by a learner that has access to decisions made by a pool of experts,
which leads to the following prediction protocol:
\makeatletter
  \renewcommand{\ALG@name}{Protocol}
\makeatother
\begin{algorithm}[H]
  \caption{Prediction with expert advice}
  \label{prot:PEA}
  \begin{algorithmic}
    \STATE $L_0:=0$.
    \STATE $L_0^k:=0$, $k=1,\ldots,K$.
    \FOR{$N=1,2,\dots$}
      \STATE Expert $k$ announces $\gamma_N^k\in\Gamma$, $k=1,\ldots,K$.
      \STATE Learner announces $\gamma_N\in\Gamma$.
      \STATE Reality announces $\omega_N\in\Omega$.
      \STATE $L_N:=L_{N-1}+\lambda(\omega_N,\gamma_N)$.
      \STATE $L_N^k:=L_{N-1}^k+\lambda(\omega_N,\gamma_N^k)$, $k=1,\ldots,K$.
    \ENDFOR
  \end{algorithmic}
\end{algorithm}
\makeatletter
  \renewcommand{\ALG@name}{Algorithm}
\makeatother
Here $L_N$ is the cumulative loss of the learner at a time step $N$, and $L_N^k$ is the cumulative loss of $k$th expert at this step. There are a lot of well-developed algorithms for the learner, probably the most known are Weighted Average Algorithm \cite{Kivinen1999}, Strong Aggregating Algorithm \cite{VovkAS,VovkGofPEA}, Weak Aggregating Algorithm \cite{KalnishkanWAAWM}, Hedge Algorithm \cite{Freund1997}, and Tracking the Best Expert \cite{HerbsterTBE}. The basic idea behind these algorithms is to assign weights to experts and then use their predictions in the correspondence with their weights in a way that minimizes the learner's loss. Weights of experts are changed at each step, which allows a prediction algorithm to adapt to the sequence of outcomes.

The Strong Aggregating Algorithm, further called the Aggregating Algorithm or the AA, has the strongest theoretical guarantees for some games with a
``sufficiently convex'' loss function, whereas the accuracy in practice
some cases can probably not be the best one. We use the Aggregating
Algorithm for the experiments described in this paper, but one can use other
online algorithms to give probability forecasts. In the case of the Brier game
with more than two outcomes only the AA and the Weighted Average Algorithm have
theoretical bounds for their losses derived in the extended arXiv version of \cite{VovkPEABG}.
The Aggregating Algorithm has a parameter $\eta$, the learning rate. It is proved that for the Brier game the best theoretical guarantees can be received if $\eta = 1$.
The theoretical bound for its cumulative loss at a prediction
step $N$ is
\begin{equation}
L_N(\mathrm{AA}) \le L_N^k + \ln K
\end{equation}
for any expert $k$, where the number of experts equals $K$. The way it makes
predictions is described as Algorithm~\ref{alg:SAA}.

\addtocounter{algorithm}{-1}
\begin{algorithm}[ht]
  \caption{Strong aggregating algorithm for the Brier game} \label{alg:SAA}
  \begin{algorithmic}
    \STATE $w_0^k:=1$, $k=1,\ldots,K$.
    \FOR{$N=1,2,\dots$}
      \STATE Read the Experts' predictions $\gamma^k_N$, $k=1,\ldots,K$.
      \STATE Set $G_N(\omega):=-\frac{1}{\eta}\ln\sum_{k=1}^K w^k_{N-1} e^{-\eta\lambda(\omega,\gamma_N^k)}$, $\omega\in\Omega$.
      \STATE Solve $\sum_{\omega\in\Omega}(s-G_N(\omega))^+=2$ in $s\in\mathbb{R}$.
      \STATE Set $\gamma_N\{\omega\}:=(s-G_N(\omega))^+/2$, $\omega\in\Omega$.
      \STATE Output prediction $\gamma_N\in\PPP(\Omega)$.
      \STATE Read observation $\omega_N$.
      \STATE $w_N^k:=w_{N-1}^ke^{-\eta\lambda(\omega_N,\gamma_N^k)}$.
    \ENDFOR
  \end{algorithmic}
\end{algorithm}
\bigskip
\section{Data set}\label{sec:data}
We are working with a data set \cite{Devetyarov2009}
that was collected over the period of 7 years
and has patients with the disease (referred to as \emph{cases})
and patients who were healthy all this period,
called
\emph{controls}.
Description of the collection process is not a goal of this
paper, so we do not state this question in detail.
More detailed description of the data set and peak extracting procedures
can be found in \cite{Menon2005} and \cite{Devetyarov2009}.
This paper develops further the analysis performed in \cite{Devetyarov2009}.

We consider prediction in \emph{triplets}.
There are 881 samples in total: 295 cases, 586 matched controls. There are up to 5 samples for each of the cases. Information for all
samples contains the value of CA125, time to diagnosis, intensities of 67 mass-
spectrometry peaks, and other. Time to diagnosis is the time interval measured
in months between the date when the measurement was taken and the date
when OC was diagnosed, or the date of operation. Peaks are ordered by their
frequency, or the percentage of samples having a non-aligned peak. We have 67
peaks of frequency more than 33\%. For classification purposes we exclude cases
with only one matched control, and cases with lack of suitable information. As a
result, we have 179 triplets containing 358 control samples and 179 case samples taken from
104 individuals. Each triplet is assigned a \emph{time-to-diagnosis} defined from the time
to the moment of diagnosis of the case sample in this triplet.

\section{Experiments}\label{sec:experiments}
This section describes two experiments.
The first is a study of probability prediction of
ovarian cancer.
The second checks that our results are not accidental by calculating $p$-values.

\subsection{Probability prediction of ovarian cancer}\label{ssec:alltr}
The aim of this experiment is to demonstrate how we give probability predictions
for samples in a triplet and compare them to predictions using CA125 only.
The outcome of each event can be represented as a vector $(1, 0, 0)$, $(0, 1, 0)$, or $(0, 0, 1)$.
The prediction of CA125 is represented as a vector $(a1, a2, a3)$. This vector is received by applying the maximum rule to CA125 levels.

We use the following procedure to construct other predictors combining
CA125 and peak intensities. For each patient we calculate values
\begin{equation}\label{eq:pred}
u(v,w,p) = v \ln C + w\ln I_p,
\end{equation}
where $C$ is the level of CA125, $I_p$ is the intensity of the $p$-th peak, $p=1,\ldots,67$,
$v \in \{0,1\}, w \in \{-2,-1,-1/2,0,1/2,1,2\}$. The total number of different
combinations, or experts, is 537: $402 = 6\times67$ for $v = 1, w \ne 0$, $134 = 2\times67$ for
$v = 0$, and $1$ for $v = 1,w = 0$. The authors of \cite{Devetyarov2009} show how such combinations can
predict cancer well up to 15 months before diagnosis.

\begin{figure}
\includegraphics[width=0.47\textwidth]{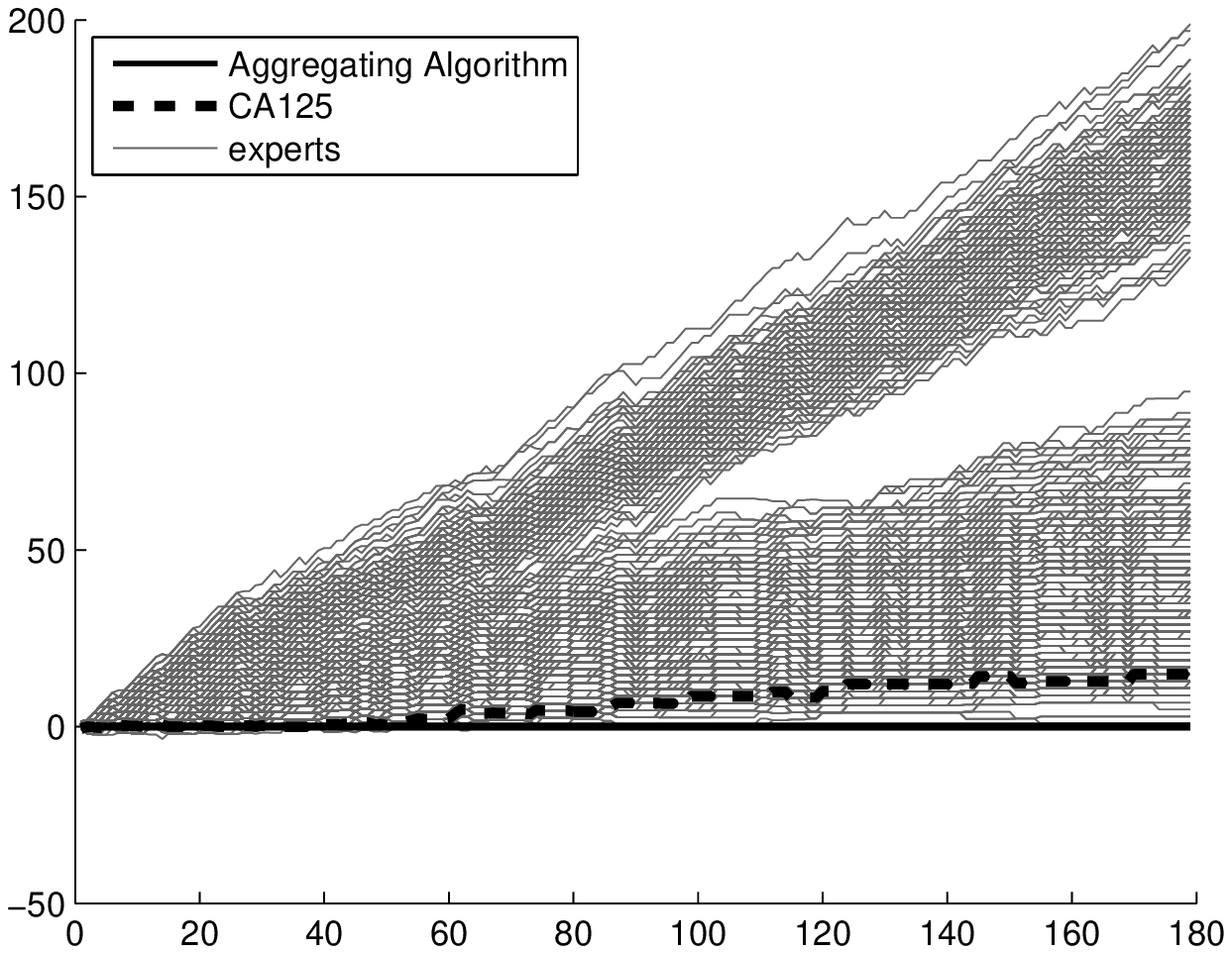} \hfill
\includegraphics[width=0.47\textwidth]{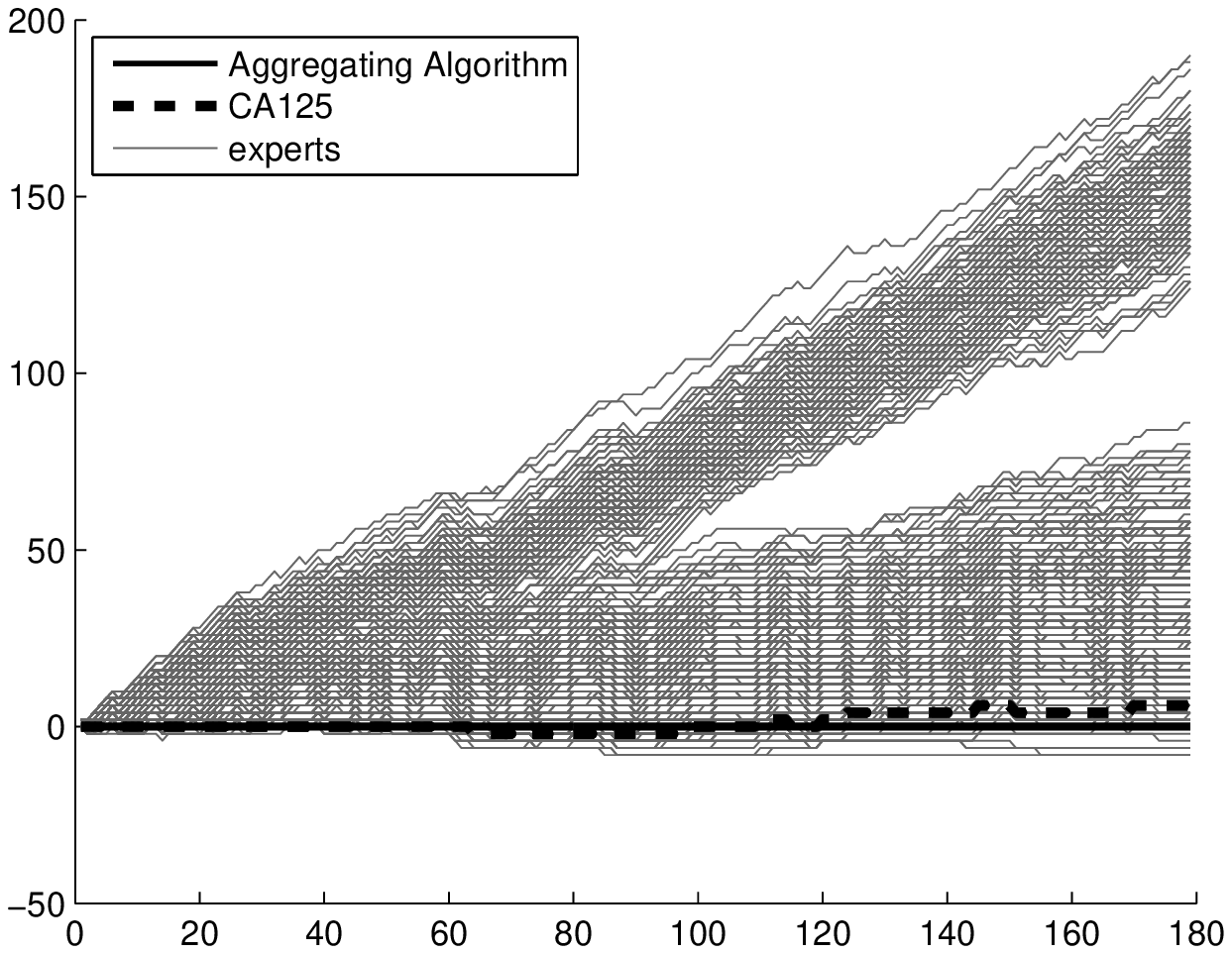} \\
\parbox[t]{0.47\textwidth}{\caption{Cumulative loss of probability predictions of the Aggregating Algorithm and other predictors over all the triplets.}\label{Fig:Prob}} \hfill
\parbox[t]{0.47\textwidth}{\caption{Cumulative loss of strict predictions of categorical AA and other predictors over all the triplets.}\label{Fig:Strict}}
\end{figure}

For online prediction purposes we sort all the triplets by the date of measurement of the case sample. At each step we give the probability of being
diseased for each person in the triplet, or numbers $p_1, p_2, p_3 \ge 0$ : $p_1+p_2+p_3 = 1$.
We choose the uniform initial distribution on the experts and the theoretically optimal value for the parameter $\eta$, $\eta=1$, of the Aggregating Algorithm. The evolution of the cumulative Brier loss of all the experts minus the cumulative loss of our algorithm over all the 179 triplets is presented in Figure~\ref{Fig:Prob}. Clearly, the line for the AA is zero
since we subtract its loss from itself. Experts having the line lower than zero are
better than the AA, experts having the line higher than zero are worse. The $x$-axis
presents triplets in the chronological order.
We can see from Figure~\ref{Fig:Prob} that the
Aggregating Algorithm predicts better than most experts in our class after about
54 triplets, in particular better than CA125. At the end the AA is better than all the
experts. The group of lines clustered on the top of the graph separated from the
main group are experts which do not include CA125. They make relatively many
mistakes especially on late stages of the disease and accumulate a large loss. This
shows that the probability predictions of the AA are more precise than predictions
of experts interpreted as probability predictions. Moreover, we can be sure that the loss of the Aggregating Algorithm will never be much worse than the loss of the best expert since there is a
theoretical bound for it \cite{VovkPEABG}.

One can say this comparison is not fair because we
allow experts give only strict predictions, and our algorithm is more flexible so
its Brier loss is not so large. On the other hand, it is not trivial to find experts
which make probability predictions, or convert CA125 to probabilities of the
disease for each sample in triplet, so this approach presents one of the ways to
generate them.

In order to make a more strict comparison we allow the AA to make only strict predictions and use the maximum rule to convert probability predictions into strict predictions. We will further refer to this algorithm as to the \emph{categorical AA}. If we calculate the Brier loss, we
get Figure~\ref{Fig:Strict}. We can see that the categorical AA still beats CA125 at the end in the case where
it gives strict predictions. The final performance is the performance on the whole
data set. In this case the loss of the categorical AA is more than the loss of some predictors.
It is useful to know specific combinations which perform well in this experiment.
At the last step the best performance is achieved by combinations
\begin{align} \label{eq:bestcomb1}
&\ln C - \ln I_{3}, \ln C - \frac{1}{2}\ln I_{3},\\
&\ln C - \ln I_{2}, \ln C - \frac{1}{2}\ln I_{7}.\notag
\end{align}
After them combinations with peaks 50, 2, 7, 1, 34, 47 follow.

\subsection{Prediction on different stages of the disease}\label{ssec:early}
Our second experiment is aimed to investigate whether it is possible to predict
better than CA125 at early stages of the disease. In this experiment we follow the approach proposed in \cite{Devetyarov2009}. We consider 6-month time intervals with starting
point $t = 0, 1,\ldots,16$ months before diagnosis. We will show further that our
predictions are not reliable for earlier stages. For each period we select only those
triplets from the corresponding time interval, the latest for each case patient if
there are more than one. We denote the number of triplets for the interval $t$ of length $\theta$
by $S_{t,\theta}$. We use $\theta = 6$.

\begin{figure}
\includegraphics[width=0.47\textwidth]{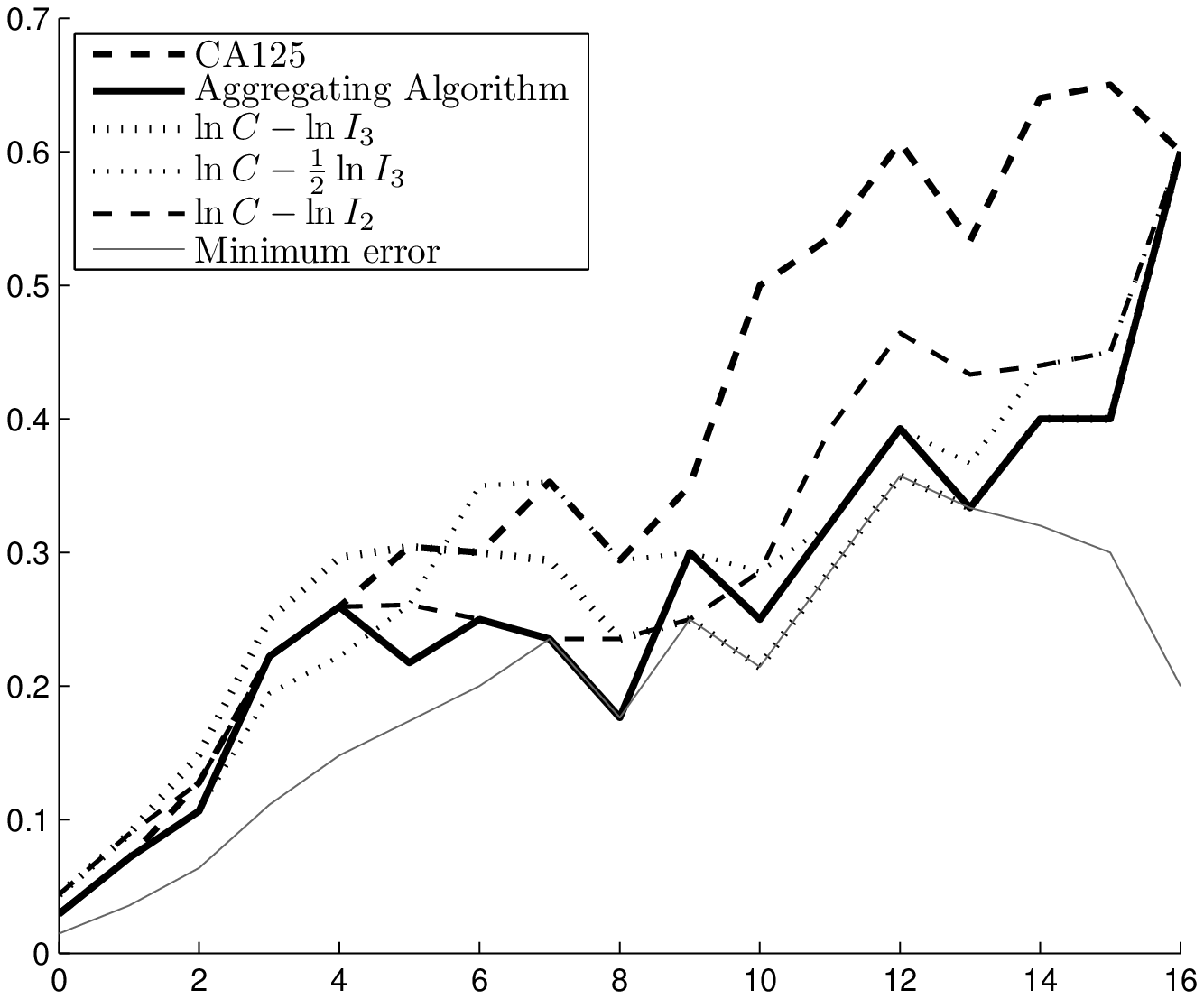} \hfill
\includegraphics[width=0.47\textwidth]{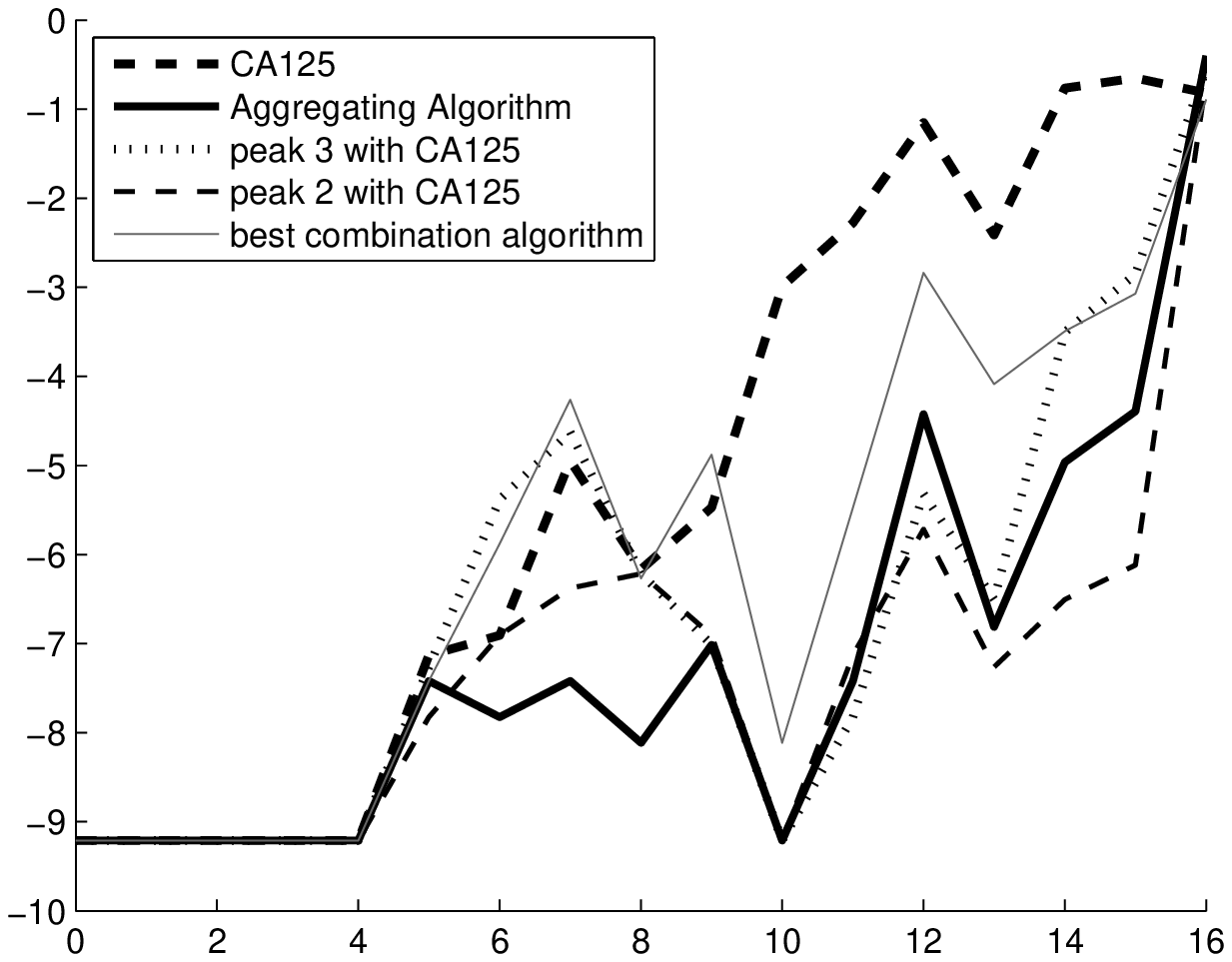} \\
\parbox[t]{0.47\textwidth}{\caption{Fraction of erroneous predictions over different time periods of different predictors}\label{Fig:Err}} \hfill
\parbox[t]{0.47\textwidth}{\caption{The logarithm of $p$-values for different algorithms}\label{Fig:pvalues}}
\end{figure}

In this experiment we do not use a uniform initial weight distribution on the
experts for the Aggregating Algorithm. Instead, we assume the importance of a peak
decreases as its number increases in accordance with a power law,
and that different combinations including the same peak have the same importance.
This makes sense because peaks are
sorted by their frequency in the data set, so peaks further down
the list are less frequent and important for fewer people.
Our specific weighting scheme is that
the combinations with peak 1 have initial weight $1 = d^0$,
the combinations with peak 2 have initial weight $d^{-1}$, etc.
We empirically choose the coefficient for this distribution
$d = 1.2$, and the parameter $\eta$ for the AA $\eta = 0.65$.
The number of errors was calculated as a half of Brier loss,
which corresponds to counting errors in the case where predictions are strict.
Figure~\ref{Fig:Err} shows the fraction of erroneous predictions made by different algorithms
over different time periods. It presents values for CA125, for the Aggregating
Algorithm, and for the best one combination of the form \eqref{eq:pred}. We
also include fractions of erroneous predictions for the three best combinations \eqref{eq:bestcomb1} as peaks 2 and 3
were noticed in \cite{Devetyarov2009} to have a good performance.

This figure shows that the performance of the Aggregating Algorithm is at least as good as the performance of CA125 on all stages before diagnosis. For
the period 9--13 months the combination $\ln C - \ln I_3$ performs better than the AA,
but on late stages 0--8 months it performs worse. Other combinations are even
worse. Thus we can say that instead of choosing one particular combination, we should use the Aggregating Algorithm to mix all the combinations. This allows
us to predict well on some stages of the disease.

The choice of the coefficients for the AA requires us
to check that our results are not accidental.
Since the amount of data we have
does not allow us to carry out reliable cross-validation procedure,
we follow the approach to calculating $p$-values proposed in \cite{Gammerman2008}.
This approach was applied for combinations \eqref{eq:pred} in \cite{Devetyarov2009}.
For each stage of the disease,
we are testing the null hypothesis that peak intensities and CA125
do not carry any information relevant for predicting labels.
Except for the earliest stages,
we prove that either this hypothesis is violated or some very unlikely event happened.

We calculate $p$-values for testing the null hypothesis.
The $p$-value can be defined as the value taken by a function $p$ satisfying
\begin{equation*}
\forall \delta \text{ Probability}(p \le \delta) \le \delta
\end{equation*}
for all $\delta\in(0,1)$ under the null hypothesis.
To calculate $p$-values we choose the test statistic $T$ described below,
apply it to our data, and get the value $T_0$.
Then we calculate the probability of the event that $T \le T_0$ under the null hypothesis.

Let $\tau$ be a triplet in $S_{t,6}$ and $\text{err}(\tau, d, \eta)$ be half loss of the categorical AA with parameter $\eta$ and initial power distribution with parameter $d$ on
the triplet $\tau$.
Then the half loss in each time interval $[t,t+6]$ is $\text{Err}(S_{t,6}; d; \eta) =
\sum_{\tau \in S_{t,6}} \text{err}(\tau; d; \eta)$, where $S_{t,6}$ is the set of triplets for the time interval $t$ $[t,t+6]$. Let us
assume that the AA with parameters $d = 1.2$ and $\eta = 0.65$
makes $N_t$ errors on the triplets from $S_{t,6}$. We randomly reassign labels in triplets.
Then for each $t$ we calculate the minimum number of errors $E$ made by the AA by the rule
\begin{equation*}
E = \min_{d \in D, \eta \in R} \text{Err}(S_{t,6}, d, \eta).
\end{equation*}
Here $D = \{1.1, 1.2,\ldots,2.0\}$ and $R = \{0.1, 0.15, 0.2,\ldots,1.0\}$, so we consider different values for all parameters of the algorithm. This number is our test statistic. The $p$-value is calculated by the Monte-Carlo procedure stated as Algorithm~\ref{alg:pvalue}.
\begin{algorithm}[ht]
  \caption{$p$-value calculation}\label{alg:pvalue}
  \begin{algorithmic}
    \STATE \textbf{Input:} $t$, time to diagnosis.
    \STATE \textbf{Input:} $N=10^4$, number of trials.
    \STATE $E_0 := \min_{d \in D, \eta \in R} \text{Err}(S_{t,6}, d, \eta)$
    \STATE $Q :=0 $
    \FOR{$j=1,\ldots,N$}
        \STATE Assign a case label to a randomly chosen sample in each triplet in $S_{t,6}$.
        \STATE Calculate $E = \min_{d \in D, \eta \in R} \text{Err}(S_{t,6},d,\eta)$ for this data set.
        \IF{$E \le E_0$}
            \STATE $Q = Q+1$
        \ENDIF
    \ENDFOR
    \STATE \textbf{Output:} $\frac{Q+1}{N+1}$ as a $p$-value.
  \end{algorithmic}
\end{algorithm}

\bigskip

The logarithms of $p$-values for different algorithms
are presented in Figure~\ref{Fig:pvalues}.
It includes values for AA.
It also includes values taken from \cite{Devetyarov2009} for the CA125 only. It
includes $p$-values for the algorithm described in \cite{Devetyarov2009}. This algorithm chooses the combination
with the best performance and the most frequent peak for each permutation of labels.
The figure also includes the $p$-values for the algorithm, which chooses the best combination with one particular peak, 2 or 3.

As we can see, our algorithm has small $p$-values, comparable with or even
smaller than $p$-values for other algorithms. But our algorithm has fewer adjustments, because it does not choose even the peak at each step, but mixes all peaks in the same manner. It does not even choose the best parameters
for every time interval but chooses them for all the time periods. The
precise values for errors and $p$-values are presented in Table~\ref{tab:erpval}. Lower index $_e$
means the half loss for a given algorithm, lower index $_p$ means the $p$-values
for a given algorithm. The $\text{Min}_e$ column shows the minimum number of errors made
by one of the combinations, the $p$ column shows the $p$-values for the method which chooses
the best combination for a current time period (see \cite{Devetyarov2009}), $C^3_{1,e}$ shows the number of errors
for the combination $\ln C -\ln I_3$, $C^3_{2,e}$ shows the number of errors for the combination $\ln C - \frac{1}{2}\ln I_3$,$C^2_{e}$ shows number of errors for the combination $\ln C -\ln I_2$.
Columns $3_p$ and $2_p$ contain the $p$-values for peaks 3 and 2 correspondingly.

\begin{table}[h]
\caption{Number of errors and $p$-values for different algorithms}\label{tab:erpval}
\begin{tabular}{|l|l|l|l|l|l|l|l|l|l|l|l|l|l|}
\hline
$t$ & $|S_{t,6}|$ & $\text{CA125}_e$ & $\text{CA125}_p$ & $\text{AA}_e$ & $\text{AA}_p$ & $\text{Min}_e$ & $p$ & $C_{1,e}^3$ & $C_{2,e}^3$ & $3_p$ & $C^2_e$ & $2_p$\\
\hline
0	&	68	&	2	&	0.0001	&	2	&	0.0001	&	1	&	0.0001	&	3	&	2	&	0.0001	&	3	&	0.0001	 \\
1	&	56	&	4	&	0.0001	&	4	&	0.0001	&	2	&	0.0001	&	5	&	4	&	0.0001	&	5	&	0.0001	 \\
2	&	47	&	6	&	0.0001	&	5	&	0.0001	&	3	&	0.0001	&	7	&	5	&	0.0001	&	6	&	0.0001	 \\
3	&	36	&	8	&	0.0001	&	8	&	0.0001	&	4	&	0.0001	&	9	&	7	&	0.0001	&	8	&	0.0001	 \\
4	&	27	&	7	&	0.0001	&	7	&	0.0001	&	4	&	0.0001	&	8	&	6	&	0.0001	&	7	&	0.0001	 \\
5	&	23	&	7	&	0.0008	&	5	&	0.0006	&	4	&	0.0006	&	7	&	6	&	0.0007	&	6	&	0.0004	 \\
6	&	20	&	6	&	0.0010	&	5	&	0.0004	&	4	&	0.0028	&	6	&	7	&	0.0046	&	5	&	0.0010	 \\
7	&	17	&	6	&	0.0071	&	4	&	0.0006	&	4	&	0.0141	&	5	&	6	&	0.0098	&	4	&	0.0017	 \\
8	&	17	&	5	&	0.0021	&	3	&	0.0003	&	3	&	0.0019	&	4	&	5	&	0.0020	&	4	&	0.0020	 \\
9	&	20	&	7	&	0.0042	&	6	&	0.0009	&	5	&	0.0076	&	5	&	6	&	0.0009	&	5	&	0.0010	 \\
10	&	28	&	14	&	0.0503	&	7	&	0.0001	&	6	&	0.0003	&	6	&	8	&	0.0001	&	8	&	0.0001	 \\
11	&	28	&	15	&	0.1028	&	9	&	0.0006	&	8	&	0.0042	&	8	&	9	&	0.0004	&	11	&	0.0008	 \\
12	&	28	&	17	&	0.3164	&	11	&	0.0120	&	10	&	0.0585	&	10	&	11	&	0.0049	&	13	&	0.0033	 \\
13	&	30	&	16	&	0.0895	&	10	&	0.0011	&	10	&	0.0168	&	10	&	11	&	0.0015	&	13	&	0.0007	 \\
14	&	25	&	16	&	0.4661	&	10	&	0.0070	&	8	&	0.0304	&	10	&	11	&	0.0301	&	11	&	0.0015	 \\
15	&	20	&	13	&	0.5211	&	8	&	0.0124	&	6	&	0.0464	&	8	&	9	&	0.0577	&	9	&	0.0022	 \\
16	&	10	&	6	&	0.4406	&	6	&	0.6708	&	2	&	0.4101	&	6	&	6	&	0.5979	&	6	&	0.5165	 \\
\hline
\end{tabular}
\end{table}
In practice, one often chooses a suitable significance level for their particular task. If
we choose it at 5\%, then we can see from the table that CA125 classification is
significant up to 9 months in advance of diagnosis (the $p$-values are less than 5\%).
At the same time, the results for peaks combinations and for AA are significant for
up to 15 months.
\section{Conclusion}\label{sec:conclusion}
Our results show that the CA125 criterion, which is a current standard for the
detection of ovarian cancer, can be outperformed, especially at early stages.
We have proposed a way to give probability predictions for the disease and
showed that predicting this way we suffer less loss than other predictors based on
the combination of CA125 and peak intensities.
We made another experiment to investigate the performance of our algorithm
at different stages before diagnosis.
We found that the Aggregating Algorithm we use to mix combinations predicts better
than almost any combination.
To check that our results are not accidental we calculate $p$-values from it under the null hypothesis that peaks and CA125 do not give any
information about the disease at a particular time before the diagnosis. Using our test statistic we get small $p$-values.
They show this hypothesis can be rejected at the standard significance level 5\% later than 16 months before diagnosis. Our test statistic produces $p$-values that are never worse than the $p$-values produced by the statistic proposed in \cite{Devetyarov2009}. There is no other papers dealing with our database. Other approaches of probability prediction of ovarian cancer using CA125 criteria based on the Risk of Ovarian Cancer algorithm (see \cite{Skates2003}) require multiple statistical assumptions about the data and a much larger size of a database. Thus they can not be comparable in our setting.

An interesting direction of future research is to consider the prediction of the probability of the disease for an individual patient, rather than put it artificially into triplets.

\section{Acknowledgments}\label{sec:acknowledgments}
We would like to thank Mike Waterfield, Ali Tiss, Celia Smith, Rainer Cramer, Alex Gentry-Maharaj, Rachel Hallett, Stephane Camuzeaux, Jeremy Ford, John Timms, Usha Menon, and Ian Jacobs for the given data set and useful discussions of experiments and results. This work has been supported by EPSRC grant EP/F002998/1 ``Practical Competitive Prediction'', EU FP7 grant ``OPTM Biomarkers'', MRC grant G0301107 ``Proteomic Analysis of the Human Serum Proteome'', ASPIDA grant ``Development of new methods of conformal prediction with application to medical diagnosis'' from the Cyprus Research Promotion Foundation, Veterinary Laboratories Agency of DEFRA grant ``Development and application of machine learning algorithms for the analysis of complex veterinary data'', and EPSRC grant EP/E000053/1 ``Machine Learning for Resource Management in Next-Generation Optical Networks''.

\end{document}